\documentclass{article} 
\usepackage{iclr2027_conference,times}

\usepackage{amsmath,amsfonts,bm}

\def\eqref#1{equation~\ref{#1}}

\def\1{\bm{1}}

\DeclareMathAlphabet{\mathsfit}{\encodingdefault}{\sfdefault}{m}{sl}
\SetMathAlphabet{\mathsfit}{bold}{\encodingdefault}{\sfdefault}{bx}{n}

\usepackage{hyperref}
\usepackage{url}
\usepackage{graphicx}
\usepackage{float}
\usepackage{booktabs} 
\usepackage[table]{xcolor}
\usepackage{fvextra}
\usepackage{mdframed}
\usepackage{soul}
\usepackage{footnote}
\makesavenoteenv{figure}

\title{CUAWright: A Minimal Unified Interface for Digital Agents}

\author{
\vspace{0.15in}
\begin{tabular}{@{}l@{}}
\textbf{
Yadong Lu$^{1}$,
Theodore Lee$^{2,\dagger}$,
Yifei Li$^{3,\dagger}$,
Lawrence Keunho Jang$^{4}$,
Tianci Xue$^{3}$,
Yu Su$^{3}$
} \\[0.05in]
\textbf{
Huan Sun$^{3}$,
Ahmed Hassan Awadallah$^{1}$
}
\end{tabular}
\\[0.1in]
{\small
$^{1}$Microsoft Research
\qquad
$^{2}$National University of Singapore
\qquad
$^{3}$The Ohio State University
}
\\[0.03in]
{\small
$^{4}$Carnegie Mellon University
}
}

\iclrfinalcopy 
\begin{document}

\maketitle

\begingroup
\renewcommand{\thefootnote}{\fnsymbol{footnote}}
\footnotetext[2]{Work done during an internship at Microsoft Research.}
\endgroup
\begin{abstract}
The prevailing approach to computer-use agents  couples a model with a domain-specific harness: a browser or desktop environment equipped with human engineered tools that are fixed before task execution.  As models’ coding capabilities improve, the GUI native and static harness prevents them from direct programmatic operation on system state, as well as flexible construction of tools. To this end, we introduce CUAWright, a minimal terminal harness of roughly 3K lines of code that uses bash commands as its sole action interface, and a file system as its evolvable space for dynamically creating tools and managing the context. We conduct comprehensive experiments across a wide range of digital tasks, and demonstrate that by giving the agent a minimal, programmable interface, it achieves substantially stronger results compared to their GUI or hybrid CLI interface across a wide range of tasks. On OSWorld 2.0, CUAWright delivers a 33.2\% relative improvement in partial reward while reducing estimated cost by 37.5\% compared with the published GPT-5.5 baseline. On Online-Mind2Web and the long horizon Odysseys benchmark, CUAWright substantially outperforms GUI native harness by 4.7\% and 44.0\% in success rate, respectively.  Furthermore, we found the gains extend to CAD applications that require accurate visual understanding and CLI interaction: on CADGenBench and BenchCAD, our unified harness yields 8.1\%-41.6\% relative improvements over other CLI based harnesses with GPT-5.5.  Together, these results suggest digital environments are far more programmable than their GUI interfaces imply, and a minimal terminal-focused harness is the key for better performance and efficiency.

\begin{figure}[ht]
\centering
\includegraphics[width=0.95\textwidth]{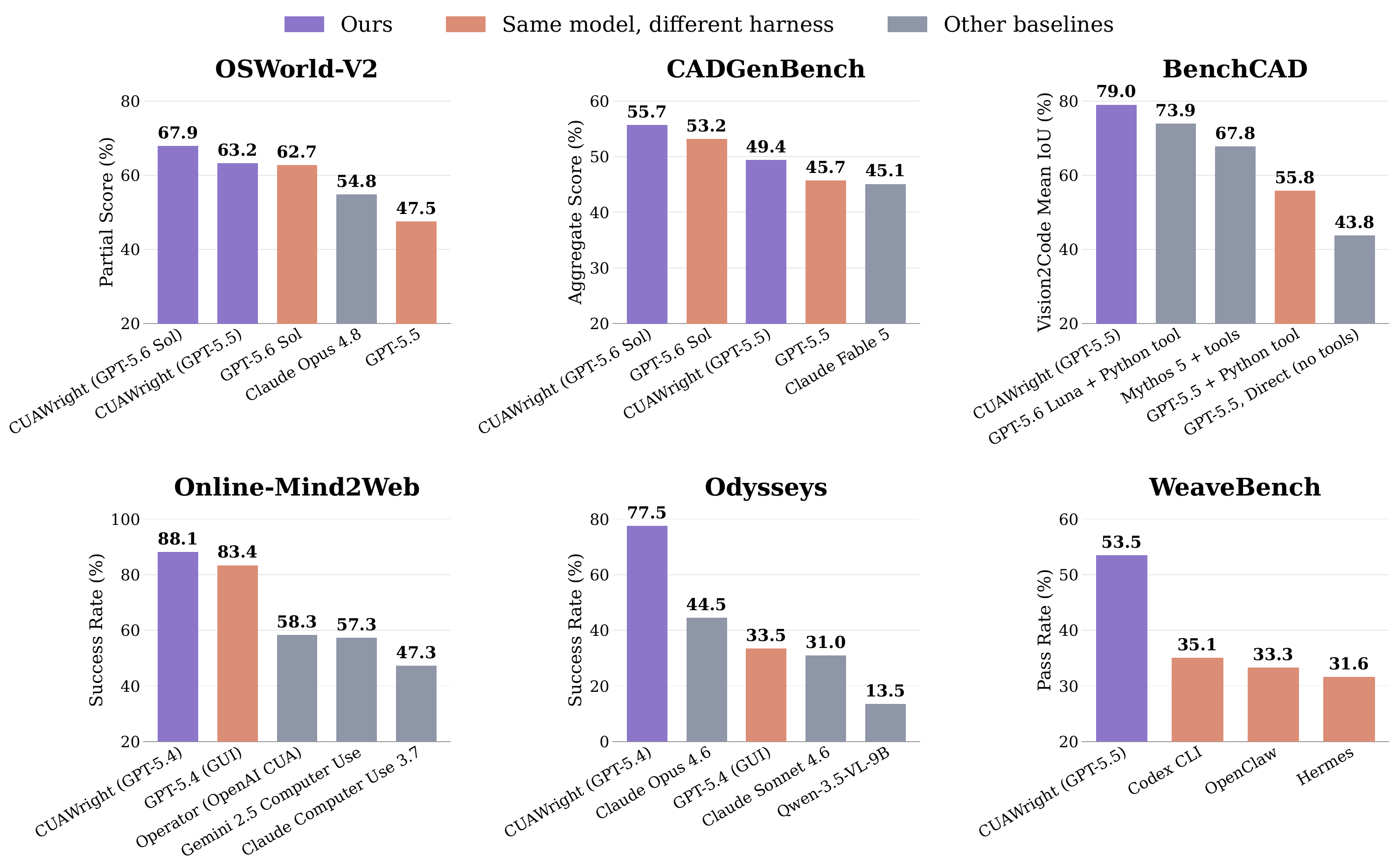}
\caption{\textbf{CUAWright} demonstrates strong performance across different digital task benchmarks. For each benchmark, we compare CUAWright with the existing GUI or hybrid CLI based approach with highest performance score using the same underlying model. CUAWright demonstrated substantially better performance compared to the baselines. 
}
\label{fig:pipeline}
\end{figure}

\end{abstract}


\section{Introduction}
\label{sec:introduction}


Computer-use agents have advanced rapidly across different domains and tasks. Web agents can observe screenshots or DOM tree of web pages at each step, and select the next elements or predict pixel coordinates to operate on directly~\citep{deng2023mind2web,zhou2024webarena, he2024webvoyager,zheng2024seeact, hong2024cogagent, pahuja2025explorer, shi2025youtuagent, gupta2026molmoweb, awadallah2026fara15}. Similarly, desktop agents can use softwares in the environments through screenshots and low-level mouse and keyboard actions like humans~\citep{xie2024osworld,bonatti2024windowsagentarena,qin2025uitars, agashe2025agents2, lu2026qwencua}.

However, solving computer-use tasks through stepwise screenshot or accessibility observations and low level actions is limiting since the screenshot exposes limited information of the environment state. And primitive actions associated with keyboard and mouse movements constrains the agent from direct and programmatic access to the underlying system state. 
More recent works explored hybrid interfaces that delegate GUI operation and code action through different sub-agents~\citep{song2025coact1,zhong2026actionengine} or explicitly define a tool to execute code against the environment~\citep{yutori2026navigator}. However, they are domain-specific and GUI centric, where agents are restricted to a fixed set of predefined tools.  
To address the above challenges, we propose a terminal native harness in which every model action is a terminal command.  As shown in Figure~\ref{fig:pipeline}, at each step the agent model predicts an action in a command line interface to engage with the underlying execution environment, which consists of a runtime, a file system and an expandable set of tools and libraries.
In particular, a browser task is driven by agent generated code written against the debugging protocol~\citep{playwright}; a desktop task is driven naturally through terminal command, CLI tools, or directly editing the source file. When a surface genuinely prevents programmatic control, the agent uses keyboard and mouse actions through xdotool/pyautogui command. 
Therefore CUAWright features: 1) a single terminal action space that is universal, allowing agent to interact with both system state and keyboard and mouse actions, and 2) direct and programmatic access to file system that the agent can use to manage its context, and craft new tools through writing, testing and executing program files using terminal commands.  
Further, in contrast to GUI native and hybrid harnesses which append a screenshot at every step, CUAWright's default observation is terminal output, and the agent captures a screenshot on demand.
We summarize our contributions into the following: 

\begin{itemize}
\item We introduce CUAWright, a minimal harness in which agents perform web and desktop tasks solely through terminal commands, and use file system for context management and new tool creation. 
\item We evaluate CUAWright across live and long-horizon web,
desktop, and computer aided design (CAD) tasks, and show that terminal-centric interaction is effective, efficient, and substantially more robust than GUI native or hybrid interaction.
\item We conduct extensive ablation studies between single agent and multi-agent settings when using terminal native harness, and demonstrate that the file system is crucial for the performance.  Detailed analysis shows emergent behaviors of frontier models crafting its own creative tools and reuse them within the task.
\end{itemize}

\section{Method}
\label{sec:method}

\begin{figure}[t]
\centering
\includegraphics[width=0.9\textwidth]{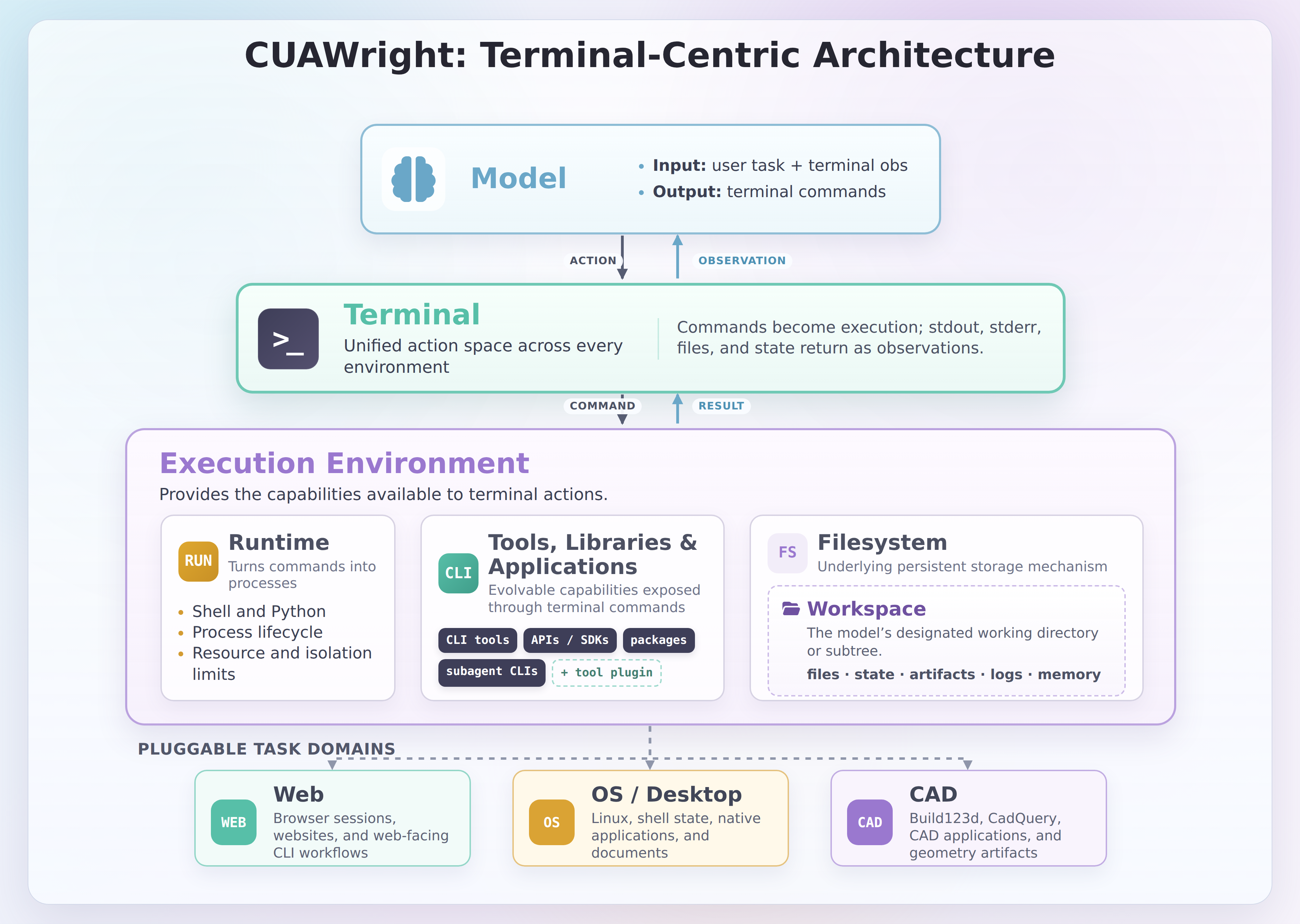}
\caption{\textbf{The CUAWright architecture.} Terminal serves as the sole interface between the agent and the task environment. At each step of task execution process, the agent model receives command line output as observation or call a CLI tool to load image into context on demand, and output commands that gets executed. The underlying execution environment contains a runtime, an expandable set of installed CLI tools and applications, and a file system as the agent's workspace to programmatically manage system state and craft new tools. }
\label{fig:pipeline}
\end{figure}


\textbf{Formulation.}
We formulate the digital agent task as a partially observable Markov decision process with environment state space $\mathcal{S}$, observation space $\mathcal{O}$, terminal action space $\mathcal{A}_{\mathrm{term}}$, and transition function $\mathcal{T}$. Given a natural-language task $T$, the agent interacts with a digital environment $\mathcal{E}$, such as a web browser, desktop application, or CAD system, to accomplish the task. At step $t$, the environment is in state $s_t \in \mathcal{S}$ and provides an observation $o_t \in \mathcal{O}$ produced by the preceding terminal execution, where $o_t$ defaults to textual output from terminal. When necessary, the agent can call a predefined CLI tool to load images into its context so that $o_t$ can also contain visual information. Conditioned on $T$, $o_t$, and the interaction history $h_t=\{(o_i,a_i)\}_{i<t}$, the agent produces an executable terminal action
\[
a_t=\pi(T,o_t,h_t), \qquad a_t\in\mathcal{A}_{\mathrm{term}}.
\]
Executing $a_t$ induces the transition $s_{t+1}=\mathcal{T}_{\mathcal{E}}(s_t,a_t)$ and produces the next observation $o_{t+1}$. We decompose the environment state as $s_t=(x_t,f_t)$, where $x_t$ denotes the underlying application and system state, and $f_t$ denotes the persistent filesystem state, including the agent workspace and artifacts created during interaction. Unlike conventional computer-use harnesses based on GUI primitives or domain-specific APIs, CUAWright{} uses executable terminal commands as a unified action interface. The action space $\mathcal{A}_{\mathrm{term}}$ remains fixed across task domains and environments, while enabling the agent to directly access and manipulate system state and dynamically construct task-specific tools.

\paragraph{Vision on demand, not by default.}
In conventional GUI-based computer-use agents, the observation typicall consists of $o_t = \{o_{t}^{screen}, o_{t}^{a11y}\}$ where $o_t^{image}$ is a screenshot at step $t$ and $o_{t}^{a11y}$ is the accessibility information such as DOM or accessibility tree. This can cause the interaction context to grow rapidly over long-horizon tasks. CUAWright instead keeps visual input optional: $o_t$ contains only the output produced by executing the terminal action $a_t$.  Only when necessary, the agent requests visual information through terminal commands in two steps: first, it calls a screenshot Python CLI tool with a specified save path; second, it calls an image-loading Python CLI tool with that path to load the captured screenshot into the agent’s context. This design substantially reduces cost compared with conventional interactive agent paradigms (e.g., roughly one-third of the cost on OSWorld~2); see Section~\ref{sec:exp:osworld} for details.

\paragraph{Persistent workspace as external state.} Let $ w_t \subseteq f_t $ denote the portion of the filesystem assigned to the agent as its local workspace. Filesystem modifications produced by terminal commands persist across interaction steps, inducing a workspace trajectory \[ w_0, w_1, \ldots, w_n. \] CUAWright introduces no separate memory state and no memory-specific interface. Instead, information that the agent wishes to retain can be externalized directly into $w_t$, including extracted values, intermediate results, task progress, logs, or records of previous failures. Memory operations are therefore ordinary terminal actions. The agent may create, read, search, append to, reorganize, or delete files using the same action space $\mathcal{A}_{\mathrm{term}}$ used to manipulate the external application. If an artifact becomes inaccurate or obsolete, the agent can modify or remove it rather than accumulating an immutable interaction history. The workspace is also independent of the model's active context window: files remain available after context compaction and can be explicitly read again when needed.

\paragraph{Dynamic tool construction in the workspace.} The same workspace also serves as a substrate for constructing new executable capabilities during task execution. Let \[ \mathcal{P}_t = \mathcal{P}_{\mathrm{env}} \cup \mathcal{P}_{\mathrm{agent}}(w_t), \] where $\mathcal{P}_{\mathrm{env}}$ contains programs initially supplied by the execution environment and $\mathcal{P}_{\mathrm{agent}}(w_t)$ contains executable artifacts created by the agent in its workspace. Through ordinary terminal actions, the agent can write a script or program into $w_t$, execute it, inspect its output, revise its implementation, and subsequently invoke it with new arguments. Once executable through the terminal, the artifact becomes part of the agent's effective program set $\mathcal{P}_t$. For example, if a task repeatedly requires an operation that is inconvenient to express through existing programs, the agent may first implement that operation as a parameterized command-line program and then reuse it in later steps.  This allows procedures discovered during interaction to be materialized as executable artifacts that can be directly tested, revised, and reused~\citep{wang2024voyager,wang2024awm,zheng2025skillweaver,wang2025inducingprogrammaticskillsagentic}.

\section{Experiments}
\label{sec:experiments}
We evaluate CUAWright on six benchmarks across web, desktop, and computer aided design (CAD), covering a wide set of high value digital task domains.


\subsection{Web Applications: Online-Mind2Web and Odysseys}
\textbf{Online-Mind2Web~\citep{xue2025illusion}} is a widely adopted benchmark to assess the performance of web agents on live real websites. It includes 300 tasks spanning 136 popular websites across diverse domains, with an LLM-as-a-Judge evaluator for assessing task completion.

We evaluated the performance of the GPT-5.4 using CUAWright on the full 300-task Online-Mind2Web benchmark. To make it compatible with the original eval settings, we enforce the agent to save critical point screenshots and to log actions during the run through prompting. We follow the official automated evaluation protocol and report the performance. 
We also compare against the GPT-5.4 baseline in a conventional screenshot-based agent setting reported in~\citep{awadallah2026fara15}, where the model predicts x,y coordinates for clicks and typing actions. Using the same underlying model, CUAWright delivers substantial improvements in medium and hard level tasks, with particularly on hard tasks, demonstrating the advantage of a code-driven, terminal-based approach over step-by-step coordinate prediction. 

\textbf{Odysseys~\citep{jang2026odysseys}} is a web agent benchmark designed to evaluate agents on realistic, long-horizon browsing tasks that involve sustained planning, memory, and cross-page reasoning spanning diverse websites. There are in total 200 long horizon tasks, where task instructions consisting of multiple atomic tasks, representing the detailed, multi-step nature of real-world web workflows. On average, 
CUAWright with GPT-5.4, achieves a 77.5\% success rate with an average of 76.1 steps, substantially outperforming the GUI based score 33.5\% using the same model. 
\begin{table}[t]
\centering
\caption{Performance comparison on two web agent benchmarks: Online Mind2Web and Odyssey.}
\label{tab:mind2web_odyssey}
\setlength{\tabcolsep}{4pt} 

\begin{tabular}{lllllllll}
\toprule
& \multicolumn{4}{c}{Online Mind2Web}
& \multicolumn{4}{c}{Odyssey} \\
\cmidrule(lr){2-5} \cmidrule(lr){6-9}
& Easy & Medium & Hard & Overall (\%) $\uparrow$
& Easy & Medium & Hard & Overall (\%) $\uparrow$ \\
\midrule

GUI baseline GPT-5.4 
& 89.9 & 81.0 & 76.6 & 83.4
& 80.0 & 58.7 & 3.7 & 33.5 \\

\rowcolor{gray!15}
CUAWright GPT-5.4
& 90.3 & 86.5 & 88.8 & 88.1 {\scriptsize [$\Delta$ +4.7]}
& \textbf{95.6} & 78.3 & 69.7 & 77.5 {\scriptsize [$\Delta$ +44.0]} \\

\rowcolor{gray!15}
CUAWright GPT-5.5
& \textbf{92.9} & \textbf{87.0} & \textbf{90.5}
& \textbf{89.4} {\scriptsize [$\Delta$ +6.0]}
& 93.3 & \textbf{89.1} & \textbf{85.3}
& \textbf{88.0} {\scriptsize [$\Delta$ +54.5]} \\

\bottomrule
\end{tabular}
\end{table}


\subsection{Desktop Applications: OSWorld 2.0}
\label{sec:exp:osworld}
On full desktop applications, we report both partial-credit scores and the stricter all-or-nothing task completion rate. All runs are evaluated on 105 of the 108 public tasks with a maximum of 300 steps. We exclude tasks \texttt{026} and \texttt{041}, which require an unavailable GitLab server, and task \texttt{082}, which requires nested Docker and is not reliably supported by our evaluation setup. As shown in Table~\ref{tab:osworld2_combined}, CUAWright achieves $63.2\%$ with GPT-5.5 xhigh and $67.9\%$ with GPT-5.6 Sol Max in the single-agent setting, significantly outperforming the corresponding official leaderboard baselines. Further, we conducted two ablation studies on the effect of single v.s. multi-agent and the effect of file system in section~\ref{ablation}.

\begin{table*}[t]
\centering
\caption{\textbf{OSWorld 2.0 desktop results and price--performance comparison.} Scores are mean partial reward; higher is better. Costs are API cost per task; lower is better. }
\label{tab:osworld2_combined}
\setlength{\tabcolsep}{4pt}
\begin{tabular}{@{}llll@{}}
\toprule
Model / harness & Partial (\%) $\uparrow$ & Strict (\%) $\uparrow$ & Cost / task (USD) $\downarrow$ \\
\midrule
\multicolumn{4}{l}{\textit{Published price--performance baselines}} \\
GPT-5.6 Luna Max & 45.57 & -- & \$7.79 \\
Claude Opus 4.8 Max & 54.80 & 20.6 & \$31.4 \\
Claude Opus 5 Max & 68.3 & 31.4 & \$28.4 \\

\midrule
\multicolumn{4}{l}{\textit{GPT-5.5 xhigh}} \\
GPT-5.5, official & 47.5 & 13.9 & \$16.1 \\
\textbf{CUAWright, multi-agent}
& 60.4 {\scriptsize ($\Delta$ +12.9)}
& 23.8 {\scriptsize ($\Delta$ +9.9)}
& \$9.6 {\scriptsize ($\Delta$ -\$6.5)} \\
\rowcolor{gray!15}
\textbf{CUAWright}
& \textbf{63.2} {\scriptsize ($\Delta$ +15.7)}
& \textbf{25.7} {\scriptsize ($\Delta$ +11.8)}
& \textbf{\$10.1} {\scriptsize ($\Delta$ -\$6.0)} \\

\midrule
\multicolumn{4}{l}{\textit{GPT-5.6 Sol max}} \\
GPT-5.6 Sol, official & 62.7 & 27.3 & \$13.9 \\
\textbf{CUAWright, multi-agent}
& 67.3 {\scriptsize ($\Delta$ +4.6)}
& 24.7 {\scriptsize ($\Delta$ -2.6)}
& \$5.3 {\scriptsize ($\Delta$ -\$8.6)} \\
\textbf{CUAWright, no file system}
& 55.5 {\scriptsize ($\Delta$ -7.2)}
& 15.5 {\scriptsize ($\Delta$ -11.8)}
& \$4.9 {\scriptsize ($\Delta$ -\$9.0)} \\
\rowcolor{gray!15}
\textbf{CUAWright}
& \textbf{67.9} {\scriptsize ($\Delta$ +5.2)}
& \textbf{27.6} {\scriptsize ($\Delta$ +0.3)}
& \textbf{\$4.4} {\scriptsize ($\Delta$ -\$9.5)} \\
\bottomrule
\end{tabular}
\vspace{2pt}
\begin{minipage}{0.98\linewidth}
\footnotesize
\end{minipage}
\end{table*}

\begin{figure}[ht]
    \centering
    \includegraphics[width=0.8\linewidth]{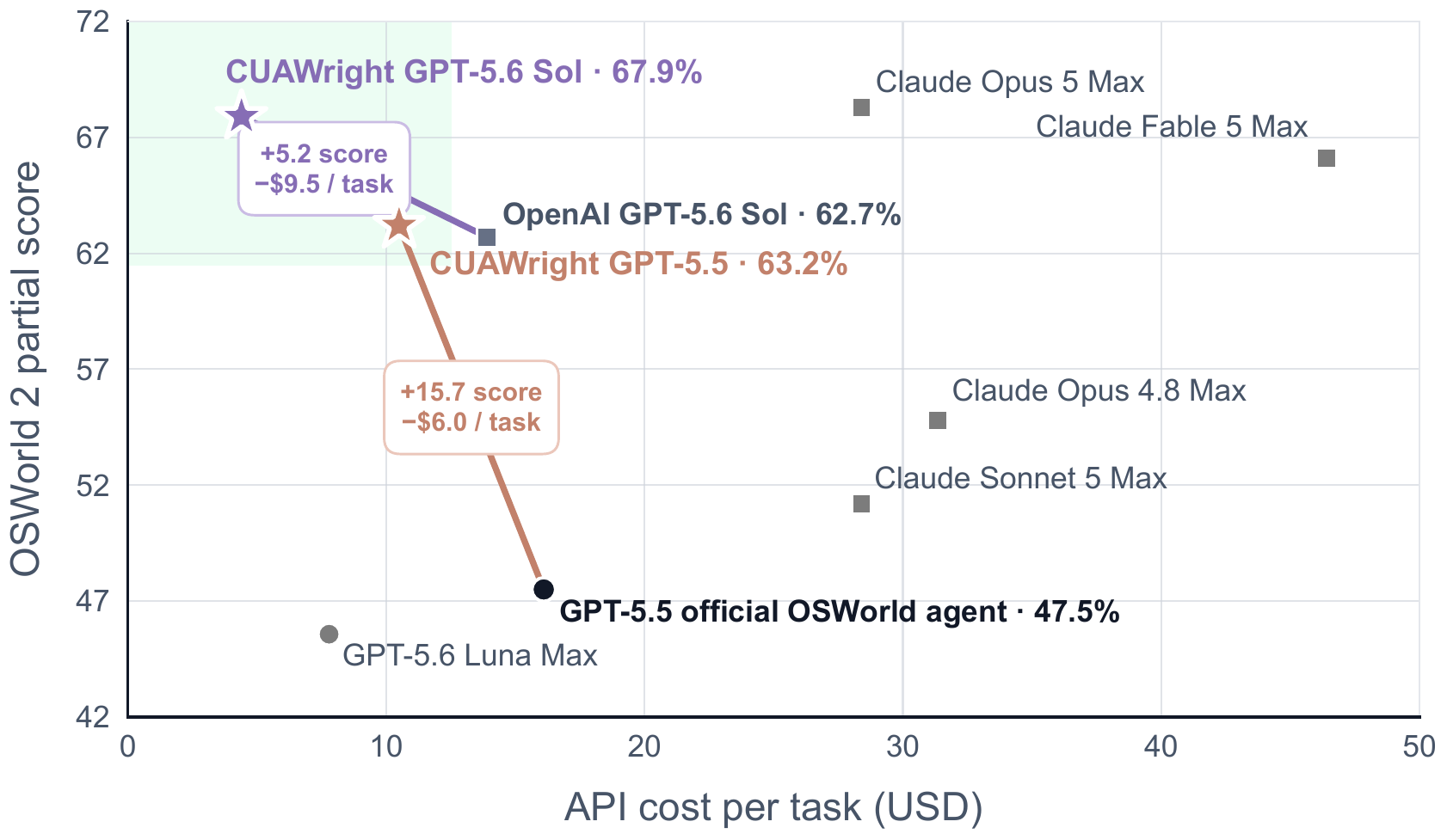}
    \caption[
        Price--performance comparison of CUAWright{} and the official
        GUI-native harness on OSWorld 2.0.
    ]{
        The price--performance comparison of CUAWright{} and the official
        GUI-native harness on OSWorld 2.0.
        CUAWright{} costs are computed from token usage cost per trajectory using public prices as of September 2026\footnotemark,
        while external baselines use their reported costs.
    }
    \label{fig:osworld-cost}
\end{figure}

CUAWright's terminal-first approach represents the frontier for performance/cost optimality. 
As shown in Figure~\ref{fig:osworld-cost}, with GPT-5.5 xhigh, CUAWright{} improves the performance by 15.7 points over the corresponding OpenAI baseline while reducing the cost by \$6.0 per task. With GPT-5.6 Sol Max, it improves the partial score by 5.2 points while reducing the per-task cost by approximately \$9.5.
 

\footnotetext{
OpenAI introduced promotional pricing for GPT-5.6 Sol on August 21, 2026,
reducing the API price from \$5/\$30 to \$4/\$20 per million
input/output tokens. Using the pre-promotion pricing, CUAWright{} with GPT-5.6 Sol costs \$5.7 per task, still substantially below the leaderboard reported cost of \$13.9 per task.
}

\subsection{Long-Horizon Desktop and Application Tasks: WeaveBench}
\label{sec:exp:weavebench}

WeaveBench complements OSWorld with 114 long-horizon tasks across eight work domains, requiring GUI observation and CLI or code execution within one trajectory. WeaveBench trajectories are then judged by an OpenClaw judge. We take the corpus and run settings from the benchmark's published GPT-5.5 high + Codex CLI configuration: the pinned 114-task dataset, high reasoning effort, and a 200-step limit.\footnote{\href{https://github.com/weavebench/WeaveBench/blob/main/docs/REPRODUCE.md}{WeaveBench reproduction settings and leaderboard.}}, and replace Codex CLI with CUAWright to evaluate its performance. Table~\ref{tab:weavebench} shows that CUAWright passes 61 of 114 tasks, improving PassRate from $35.1\%$ to $53.5\%$ and substantially improves the overall score from $49.9$ to $66.4$ compared to the Codex harness. Compared to all the other harnesses that combine the bash tool with predefined benchmark specific GUI interaction tools, our result implies that a single terminal native tool is sufficient to achieve high performance.
\begin{table}[ht]
\centering
\caption{\textbf{WeaveBench results} on the full 114-task suite using GPT-5.5 at high reasoning effort and a 200-step agent limit. PassRate is the fraction of tasks with score at least $0.80$; Overall is the mean per-task score.}
\label{tab:weavebench}
\small
\begin{tabular}{@{}llrr@{}}
\toprule
Harness & Agent & PassRate (\%) $\uparrow$ & Overall $\uparrow$ \\
\midrule
Codex CLI & GPT-5.5  & 35.1 & 49.9 \\
OpenClaw & GPT-5.5  & 33.3 & 46.6 \\
Hermes & GPT-5.5  & 31.6 & 46.6 \\
\rowcolor{gray!15}
\textbf{CUAWright} & \textbf{GPT-5.5 } & \textbf{53.5} & \textbf{66.4} \\
\bottomrule
\end{tabular}
\end{table}

\subsection{Computer-Aided Design from Images: BenchCAD}
\label{sec:exp:benchcad}

BenchCAD evaluates an agent's ability to reconstruct a 3D solid from four orthographic views by generating a CadQuery program. The resulting geometry is compared with the ground-truth solid using voxel intersection-over-union. We report the IoU score, computed as voxel IoU multiplied by the execution success rate, with non-executable outputs receiving a score of 0. We use BenchCAD's Vision2Code mode, which provides the evaluated model with a Python sandbox to render, measure, and iteratively refine its reconstruction before submission.  
Due to computational budget constraints, we evaluate on a subset of 212 samples uniformly sampled across the benchmark-provided difficulty levels (easy, medium, hard) from the full set of 17,900 samples, comprising 72 easy, 72 medium, and 68 hard tasks.
As shown in Table~\ref{tab:benchcad}, OpenAI reported both single turn performance (no tool), and multi turn performance (with benchmark provided tool).

In the with tool setting, CUAWright reaches a mean IoU of $79.0\%$ substantially increased from the $55.8\%$ of the OpenAI reported numbers on the leaderboard with GPT-5.5. 
For the no tool setting, since it is a single turn response by the model, it is harness independent.  Even though the evaluation subset for the OpenAI reported numbers is not specified and different from ours, we see our reproduced no tool score 43.8\% is slightly lower to the reported numbers 44.4\%, suggesting that our randomly selected tasks are comparable in difficulty or slightly more challenging. Despite evaluating on a harder subset, CUAWright still scores 23.2\% higher than reported GPT-5.5 numbers (with tool). 

\begin{table}[ht]
\centering
\caption{\textbf{BenchCAD Vision2Code results} across different difficulty levels; higher is better. "No tool" mode tests the agents ability to output the CAD artifacts in a single agent turn. “With tool” mode provides the model a Python sandbox with the task images and CAD libraries, allowing it to render, measure, and iteratively refine its CAD program before submitting the final answer. The official leaderboard does not provide the breakdown scores for different difficulty levels, therefore we left it to be empty.}
\label{tab:benchcad}
\small
\begin{tabular}{lllll}
\hline
& \multicolumn{1}{l}{IoU (\%) $\uparrow$} 
& \multicolumn{3}{l}{Voxel IoU by difficulty (\%) $\uparrow$} \\
\hline
& Mean  & Easy & Medium & Hard \\
\hline

\multicolumn{5}{l}{\textit{no tool}} \\
GPT-5.5, official score
    & 44.4 & -- & -- & -- \\
\rowcolor{gray!15}
GPT-5.5, ours reproduced
    & 43.8 {\scriptsize ($\Delta -0.6$)}
    & 43.4
    & 44.6
    & 43.3 \\

\hline
\multicolumn{5}{l}{\textit{with tool}} \\
GPT-5.5, official score
    & 55.8 & -- & -- & -- \\
\rowcolor{gray!15}
\textbf{GPT-5.5 CUAWright}
    & \textbf{79.0} {\scriptsize ($\Delta +23.2$)}
    & \textbf{78.6}
    & \textbf{81.4}
    & \textbf{76.9} \\

\hline
\end{tabular}
\end{table}



\subsection{Computer-Aided Design with a Tool Server: CadGenBench}
\label{sec:exp:cadgen}

CadGenBench evaluates how well a model can turn a description of a mechanical part iteratively into a valid, geometrically correct 3D model. It has a private hosted evaluator on HuggingFace to prevent benchmark contamination. We compare the codex harness with CUAWright, both of them leverage the same computer-aided design kernel. The difference is that the baseline receives its operations as registered Build123d-MCP tools, whereas CUAWright receives no CAD-specific API tool and must invoke the same MCP kernel as ordinary shell commands. 
Table~\ref{tab:cadgen} reports the comparison on all 81 public fixtures. Overall score improves by $3.7\%$ for GPT-5.5 and $2.5\%$ for GPT-5.6 Sol.
Note that all of the reported scores are produced by the privately hosted ground truth evaluator which ensures no ground truth answer leakage could happen. This result suggests the advantage of a simple agent terminal native harness over a general multiagent harness like Codex. 

\begin{table}[ht]
\centering
\caption{\textbf{CadGenBench results} on all 81 public fixtures.
We report the validated baseline score from the official benchmark using Codex with registered Build123d MCP tools. CUAWright uses the Build123d package
through CLI calls in the terminal. }
\label{tab:cadgen}
\small
\begin{tabular}{llll}
\toprule
 & Overall (\%) $\uparrow$ & Generation (\%) $\uparrow$ & Editing (\%) $\uparrow$ \\
\midrule

\multicolumn{4}{l}{\textit{GPT-5.5}} \\
Codex + Build123d-MCP tools
    & 45.7 & 37.2 & 58.8 \\
\rowcolor{gray!15}
CUAWright
    & \textbf{49.4} {\small ($\Delta$ +3.7)}
    & \textbf{40.2} {\small ($\Delta$ +3.0)}
    & \textbf{63.5} {\small ($\Delta$ +4.7)} \\

\hline

\multicolumn{4}{l}{\textit{GPT-5.6}} \\
Codex + Build123d-MCP tools
    & 53.2 & 45.1 & 65.6 \\
\rowcolor{gray!15}
CUAWright
    & \textbf{55.7} {\small ($\Delta$ +2.5)}
    & \textbf{49.0} {\small ($\Delta$ +3.9)}
    & \textbf{66.0} {\small ($\Delta$ +0.4)} \\

\bottomrule
\end{tabular}
\end{table}

\subsection{Analysis}
\subsubsection{Ablation Study}
\label{ablation}
\textcircled{\scriptsize 1} \textbf{Ablation 1: Multiple v.s. Single agents.} 
Previous studies have shown benefits of multi-agent architecture for GUI native and hybrid interfaces~\citep{song2025coact1,yang2026stateact}, we study the effect of multi-agent in terminal native CUAWright harness. 
In the docker environment, $\mathcal{P}_{\mathrm{env}}$ we install 2 new independent sub agents as CLI tool that the main terminal agent can call: 1) a planner agent, instructed to make a plan and create criteria for the self-verification before submission, 2) a self-verification agent, instructed to verify whether the task is completed given the traces and the criteria provided by the planner agent. 

The results in table~\ref{tab:osworld2_combined} show that single agent performs as well as the multiagent settings. We show two representative examples in Appendix~\ref{sec:appendix:trace:orchestration} demonstrating one case here multi-agent helps and another where single agent is better. Through careful manual review of the agent traces, we found that for OS World tasks, the planner agent often struggles to create accurate success criteria for the agent to self verify its own answer before submission. And we found that frontier models like GPT-5.5 and GPT-5.6 Sol already learns the behavior of verifying its own answer even without a standalone verifier agent. 
In addition, we find that the multi-agent setting substantially increases the median number of task turns: from 101 to 151 for GPT-5.6-sol, and from 73 to 137 for GPT-5.5.


\textcircled{\scriptsize 2} \textbf{Ablation 2: With v.s. without file system.}
As the file system is a core component of CUAWright{}, enabling the agent not only to directly manipulate files, but also to maintain persistent memory and create reusable tools during execution, we conduct an ablation study to assess its contribution. Specifically, we explicitly enforce no direct programmatic file system operations such as checking, editing and creating files using terminal command through prompting (exact prompt used can be found in Appendix~\ref{appendix_filesys_prompt}). 
The agent can still modify files indirectly through GUI operations using mouse and keyboard actions with xdotool/pyautogui CLI tool.
As shown in the table~\ref{tab:osworld2_combined}, restricting the agent from programmatic access to file system leads to a significant decrease in the performance of GPT 5.6 Sol (-12.1 pts). This suggests that programmatic control is essential to terminal-native harness.

\subsubsection{Qualitative Analysis}
\textcircled{\scriptsize 3} \textbf{Can agents craft their own tools during task execution to improve efficiency?} 
One benefit of CUAWright is that it offers a local workspace where the agent can create tools during task execution, and reuse them. In OSWorld 2 evaluation, we found in 89 (84.8\%) tasks the GPT-5.6 Sol agent explicitly created at least one saved code or executable-like file. In total there are 776 executable files created, where 192 (24.7\%) were reused (appeared in at least two distinct execution-command events). The created tools focus on inspection and extraction, interface automation, Artifact construction/editing and verification/testing.  

The figure~\ref{fig:tool_craft} in the Appendix shows for task 005, the user needs to go through the files on Desktop and find the medical records related to 2024 medical visits and  personal information. The agent first crafts a new tool to group all image files into a single composite image, and extracts all needed information at once, instead of individually check each image to see if it is related to medical information and gather results, which is much more efficient than GUI based approach.

\textcircled{\scriptsize 4} \textbf{How far can agents go without relying on primitive actions?}

We classify a task as Terminal-only when the actions contain no xdotool, pyautogui, or equivalent operating-system input invocation. And found that tasks that are terminal only performs significantly better than mixed terminal and GUI operation tasks. For examples, Terminal-only includes Conda/pip setup and browser automation without OS input. Mixed includes dinosaur-game keypresses, puzzle drags, and traces that use xdotool only to focus or verify an application.

\begin{table}[h]                                                                 \centering
\caption{\textbf{OSWorld 2.0 strict terminal-only versus mixed execution for the GPT-5.6 Sol Max} Terminal-only traces contain no xdotool, pyautogui, or equivalent operating-system input invocation. }
\label{tab:osworld2_terminal_mixed}
\small
\begin{tabular}{@{}lrrr@{}}                                                       \toprule
Execution mode & Tasks, $n$ (\%) & Mean partial (\%) $\uparrow$ & Exact completion (\%) $\uparrow$ \\
\midrule                                                                          Terminal-only & 30 (28.6) & 71.89 & 33.3 {\scriptsize (10/30)} \\                 Mixed terminal + GUI & 75 (71.4) & 66.32 & 25.3 {\scriptsize (19/75)} \\
\midrule
\rowcolor{gray!15}
\textbf{Overall} & \textbf{105 (100.0)} & \textbf{67.91} & \textbf{27.6} {\scriptsize \textbf{(29/105)}} \\
\bottomrule                                                                       \end{tabular}
\end{table}
\textcircled{\scriptsize 5} \textbf{Failure-mode analysis}

\begin{figure}[ht]
    \centering
    \includegraphics[width=0.8\linewidth]{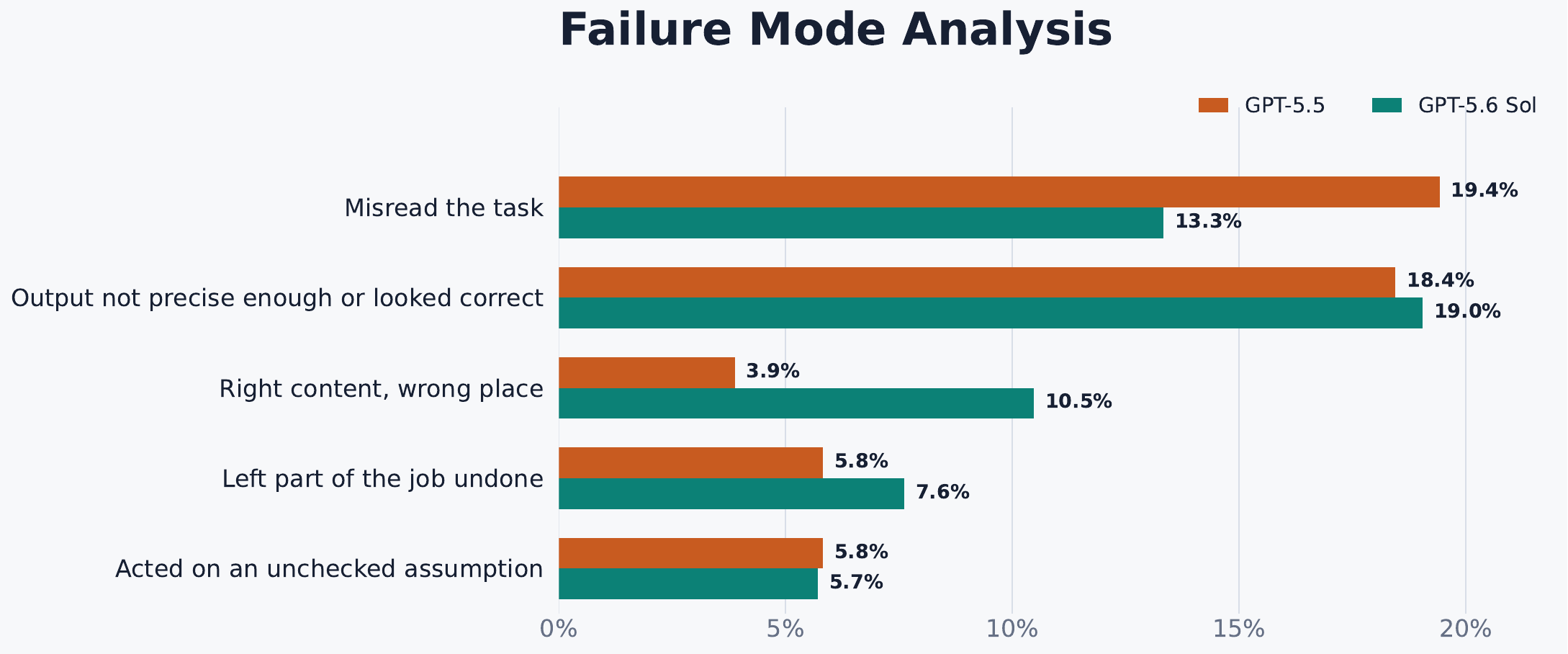}
    \caption{Top 5 failure reasons of GPT-5.5 and GPT-5.6 Sol run on OSWorld 2.0.
    }
    \label{fig:failure}
\end{figure}
The failure-mode analysis reveals distinct error profiles across the two models\ref{fig:failure}. GPT-5.5 most frequently misread the task (19.4\%), whereas GPT-5.6 Sol was more affected by outputs that appeared correct but lacked sufficient precision (19.0\%). GPT-5.6 Sol also more often produced the right content in the wrong location (10.5\% versus 3.9\%) or left part of the task incomplete (7.6\% versus 5.8\%). Unchecked assumptions occurred at nearly identical rates for both models (about 5.8\%).

\section{Related Work}
\label{sec:related_work}


\paragraph{Web and desktop interaction.}
Early web agent interacts with the browser through structured elements such as DOM~\citep{yao2022webshop, deng2023mind2web, zhou2024webarena,le2025browsergym}, with screenshot as visual context~\citep{koh2024visualwebarena,drouin2024workarena,he2024webvoyager,jang2026odysseys}, while WebVoyager and SeeAct ground multimodal decisions on live websites~\citep{he2024webvoyager,zheng2024seeact}. While for desktop agents, majority of works treat mouse, and keyboard as a universal interface~\citep{zheng2024seeact, shaw2023pixels,qin2025uitars, xie2024osworld,bonatti2024windowsagentarena,wang2026mmbenchgui,yuan2026osworld2,jang2026mypcbenchbenchmarkpersonallyintelligent}. 
More recently, Qwen-UI-Agent unifies GUI actions with CLI execution and batched actions across mobile, desktop, and web environments ~\citep{zhou2026qwenuiagent}.
However, these interaction paradigms remain tied either to environment specific representations or repeated low-level visual grounding. Also in~\cite{yuan2026osworld2}, the authors found that recent frontier models increasingly favor programmatic actions even in GUI-native action space. Our design takes a step further: we set terminal command to be the only primary model action, and GUI control is invoked programmatically, and used as a fallback rather than being the dominant action space.

\paragraph{Code-first and hybrid alternatives.}
The closest work increasingly acts on program state rather than pixels. CoAct-1 delegates between a GUI operator and a programmer agent, reducing brittle interaction chains through Python and Bash actions ~\citep{song2025coact1}.  Screenshots or Tools? studies GUI--MCP routing and finds that providing programmatic tools does not ensure that agents will adopt them effectively~\citep{fan2026screenshots}.  StateAct makes program state primary, uses GUI specialists only where needed, and verifies persisted outputs with an independent completion gate~\citep{yang2026stateact}.  StructAgent addresses a complementary problem by maintaining a compact, verifier-backed causal representation of progress for checkpointing and recovery ~\citep{wu2026structagent}.  Most directly, Terminal Agents Suffice shows that a terminal-and-filesystem coding agent can outperform GUI- and MCP-based alternatives on enterprise platforms by writing code against their APIs ~\citep{bechard2026terminalagents}.  Our work shares this terminal-first motivation but studies a broader claim: whether one command action space and filesystem context can support live-web, desktop, and graphical application tasks, including settings that are not reducible to enterprise APIs.

\section{Conclusion}
\label{sec:conclusion}

We introduced CUAWright, a minimal harness that gives the agent bash as its sole action interface and the file system as a workspace for directly operating on system state and building new tools. We conducted comprehensive experiments on six benchmarks covering a wide range of digital tasks including web, desktop and CAD applications. CUAWright demonstrated that a universal terminal interface improves both the performance and cost efficiency of computer-use agents compared with GUI-only and hybrid interfaces supported by domain-specific harnesses. As frontier LLMs rapidly advance in coding and vision, we expect computer-use agents to increasingly benefit from a unified, terminal-only interface that enables seamless operation across code and low level action spaces.

\section{Acknowdgement}

We thank Abhishek Goswami, Zach Nussbaum, Hussein Mozannar, Saaket Agash, Lingrui Xu, Chao Huang, Yelong Shen and others for feedbacks and helpful discussions.

\section*{AI Use Statement}

We used AI to assist with routine software engineering for the experimental harness and occasionally experiment execution. All AI-assisted work was reviewed and verified by the authors. We take responsibility for the final content of this work.

\bibliography{iclr2027_conference}
\bibliographystyle{iclr2027_conference}

\appendix
\section{Appendix}

\subsection{Scope and Limitations}
\label{sec:exp:scope}
One important limitation of CUAWright is that directly modifying the underlying system state may introduce additional risks. Agent actions that change files, configurations, application state, or other persistent resources can have unintended or irreversible side effects, especially when the agent operates with incomplete observations or makes an incorrect decision. Such modifications also raise practical questions around permission control, state consistency, rollback, and recovery from failed actions. In this work, we therefore focus primarily on interaction and execution under controlled settings, and leave robust mechanisms for safe state modification—such as transactional execution, validation, sandboxing, and automatic rollback—to future work.

\subsection{Implementation Details}
\label{sec:appendix:impl}

For web benchmarks Online-Mind2Web and Odysseys, CUAWright runs its workspace on the host and executes each command with the
system shell under a per-command timeout, with interactive pagers disabled.
Observations are truncated at twenty-four thousand characters, the transcript is
compacted every twenty steps, and a run is capped at one hundred steps. 

For desktop tasks, CUAWright runs each task in a container-hosted Ubuntu virtual machine provided
by the OSWorld 2.0 environment~\citep{yuan2026osworld2}, with hardware
acceleration where available, and reaches it through the guest execution
endpoint. Commands are capped at three hundred seconds and their output at
sixteen thousand characters, retained as a head and a tail around an explicit
truncation notice so that both the beginning and the end of a long output
survive. A run is capped at three hundred model steps in total across all roles. Final detailed cost for CUAWright single agent settings is in Table~\ref{tab:osworld_cost_stats}.


\begin{table}[h]
\centering
\caption{OSWorld 2.0 evaluation statistics and cost comparison.}
\label{tab:osworld_cost_stats}
\setlength{\tabcolsep}{10pt}
\small

\begin{tabular}{lll}
\toprule
\textbf{Metric} & \textbf{GPT-5.5 xhigh} & \textbf{GPT-5.6 Sol max} \\
\midrule
Tasks scored          & 105/105      & 105/105      \\
Partial score         & \textbf{63.2242\%} & \textbf{67.9130\%} \\
Exact passes          & 27/105       & 29/105       \\
Input tokens          & 400,638,827  & 434,755,823  \\
Output tokens         & 14,967,422   & 4,394,423    \\
Total tokens          & \textbf{415,606,249} & \textbf{439,150,246} \\
Current Standard cost & \textbf{\$1,054.93} & \textbf{\$458.93} \\
Cost per task         & \textbf{\$10.05} & \textbf{\$4.37} \\
\bottomrule
\end{tabular}
\end{table}




\subsection{Prompt for Ablation Study of Filesystem}
\label{appendix_filesys_prompt}
Below is the prompt we used to enforce no filesystem related operations. 

\begin{mdframed}[
    backgroundcolor=gray!10,
    linecolor=gray!10,
    linewidth=0pt,
    roundcorner=2pt,
    innerleftmargin=8pt,
    innerrightmargin=8pt,
    innertopmargin=6pt,
    innerbottommargin=6pt
]
\begin{Verbatim}[breaklines=true, breakanywhere=true, fontsize=\small, breaksymbolleft={}]
You are Actor in an Ubuntu desktop VM. Complete the task, verify the result, and submit.

Treat the task instruction as the central, top-level requirement. Follow its exact wording, named application, requested result, and restrictions.

First, re-read the instruction. As a general guide, inventory whichever parts of the task-relevant environment are useful, such as open applications, documents, browser tabs, accounts, and files in locations like Desktop, Downloads, and Documents. Let the task context determine the method and scope. Inspect any source or reference that must be matched. If a required file or fact is missing, ask the user one concise question.

Use run_command for commands, automation, application control, and inspection.

Filesystem restriction:
- Do not use run_command to perform filesystem writes or changes directly. Before running a command, consider whether it would change filesystem state.
- Do not use cat or sed with redirection, shell editing tools, Python file APIs, or similar command-line strategies to make filesystem changes. Read-only inspection and in-memory or stdout analysis remain allowed.
- Make required changes through the named application's GUI or other user-facing controls, using xdotool, CDP, or accessibility as appropriate. Normal Open, Save, Save As, Import, Export, and browser-download actions triggered through the GUI are allowed.
- A temporary screenshot created solely for the immediately following image_show.py call is allowed.

- Reuse the application, browser instance, profile, document, and state opened by setup.
- Use the application and output format named in the task.
- Complete every explicit requirement and avoid unrelated changes.
- Prefer minimal edits and exact typed values.
- Save, export, send, upload, or download to the exact requested destination.
- For GIMP, do not save or export unless explicitly requested.

Verify the final result directly:
- visual result: inspect a full-size screenshot;
- exact value or property: read it back;
- file or document: reopen the saved artifact;
- web action: reload and confirm the resulting record;
- behavior: exercise it and confirm the state change.

If verification finds a defect, repair it and verify again. Stop when the task is complete.

Tools:
- Image: `python /opt/cuawright-tools/image_show.py --path ABSOLUTE_PATH`
- Ask user: `python /opt/cuawright-tools/ask_user.py --question "TEXT"`
- Submit: `python /opt/cuawright-tools/submit.py`

Run a `/opt/cuawright-tools/*.py` tool as the only command in its run_command call.

Browser:
- Use the existing Chrome instance and profile.
- Connect at `http://127.0.0.1:1337` with `/home/user/.local/share/osw/venv/bin/python`.
- Prefer `Accessibility.getFullAXTree` to inspect controls and state.
- Use CDP, xdotool, screenshots, keyboard, and mouse to operate user-facing controls.

When complete and verified, run submit.py.
\end{Verbatim}
\end{mdframed}

\subsection{Qualitative Examples of CUAWright on different benchmarks}
\begin{figure}[ht]
    \centering
    \includegraphics[width=0.95\linewidth]{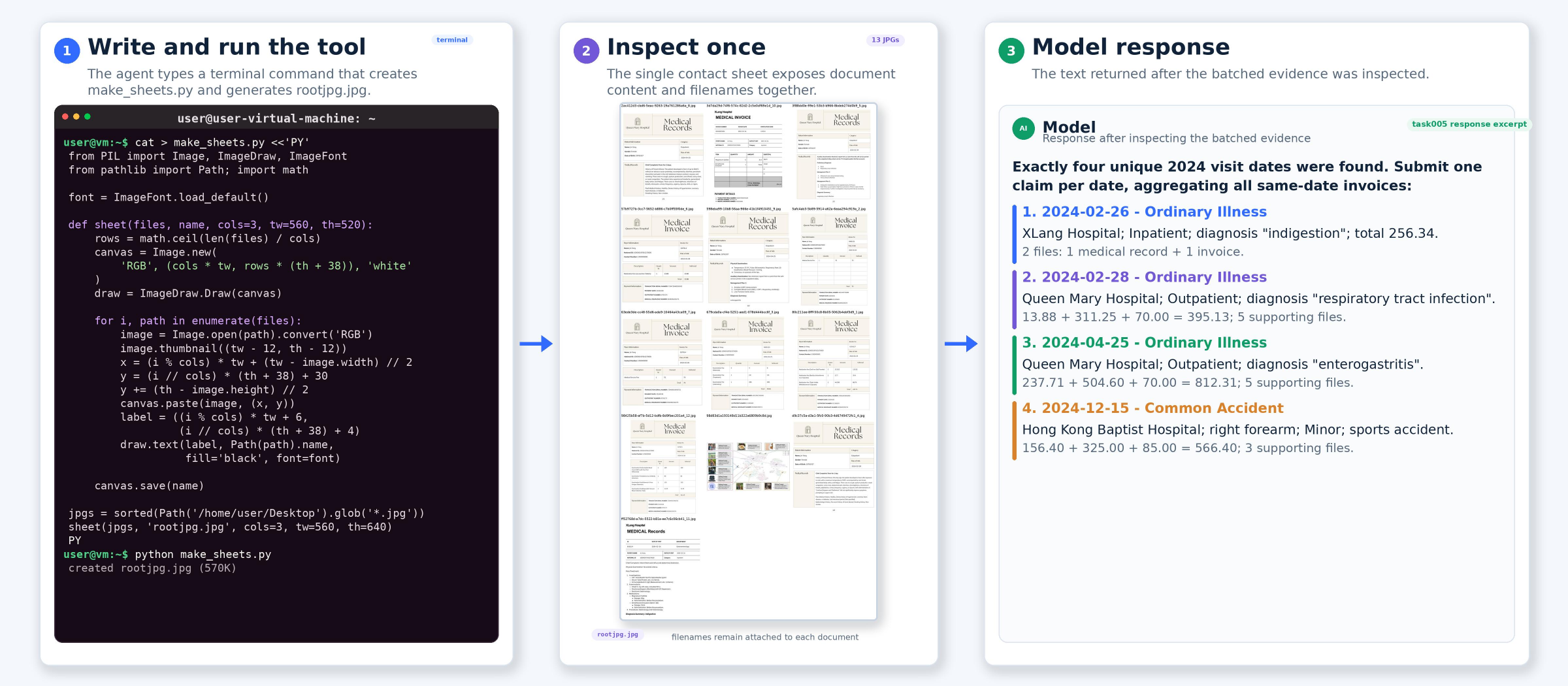}
    \caption{ Tool-crafted batched visual inspection in Task005. The agent writes and executes a lightweight sheet() utility in the terminal (left), combining 13 JPEG documents into one filename-annotated contact sheet (center). A single multimodal inspection replaces repeated GUI-based image opening and produces a structured response grouping the medical evidence by visit date, diagnosis, expenses, and supporting files (right).}
    \label{fig:tool_craft}
\end{figure}

\subsubsection{WeaveBench: Diagnosing a Checkout Latency Incident}
\label{sec:appendix:trace:weavebench}

The following trace shows the same mechanism applied across a broader systems
diagnosis. In WeaveBench task
\texttt{DAV\_task\_17\_jaeger\_trace\_root\_cause\_compare}, the agent was
asked to identify the source of a checkout latency incident and leave a
reproducible evidence bundle. The task required raw Jaeger exports, a latency
distribution, a slow trace and its longest leaf span, an OTLP packet capture,
four native Jaeger screenshots, a hypothesis, and a root-cause report.

The first API request failed because port 16685 spoke HTTP/0.9 rather than the
expected Jaeger HTTP API. The agent probed the neighboring service, found the
working endpoint on port 16686, and continued from observed state rather than
repeating the original assumption. It then queried the service and operation
indexes, generated additional real traces with the provided emitter, exported
190 \texttt{POST /checkout} traces, reconstructed their parent--child
relationships, and computed the root-span distribution. The resulting median
was 116.857 ms, while p95 and p99 were 3361.620 ms and 3558.904 ms,
respectively.







The agent did not stop at the programmatic diagnosis. It used \texttt{tcpdump}
on the collector port while replaying the real emitter, launched Chrome against
the Jaeger UI, navigated to the selected slow trace, expanded the bottleneck
span, and opened a fast-versus-slow comparison. It inspected each screenshot
before submission and wrote a report that separated detection from attribution:
RED metrics established that checkout was slow, while the trace waterfall
localized the delay to the downstream payments gateway. The report recommended
a fail-fast timeout, circuit breaking around repeated 504 responses, controlled
fallback behavior, dependency ownership, and trace-based regression gates.

The trajectory used 45 model calls and 45 terminal commands over approximately
26.6 minutes. It received a final WeaveBench score of \textbf{0.93875}.
Machine-readable exports, the packet capture, and the written diagnosis were
fully correct. The remaining deductions were mostly presentational: the search
screenshot did not isolate a perfectly clean slow cluster, and the comparison
view was a merged difference graph rather than the evaluator's preferred strict
side-by-side layout. This example exercises environment discovery, API use,
statistical analysis, packet capture, browser automation, visual inspection,
causal diagnosis, and technical communication within one trajectory.

\begin{figure}[ht]
    \centering
    \includegraphics[width=0.95\linewidth]{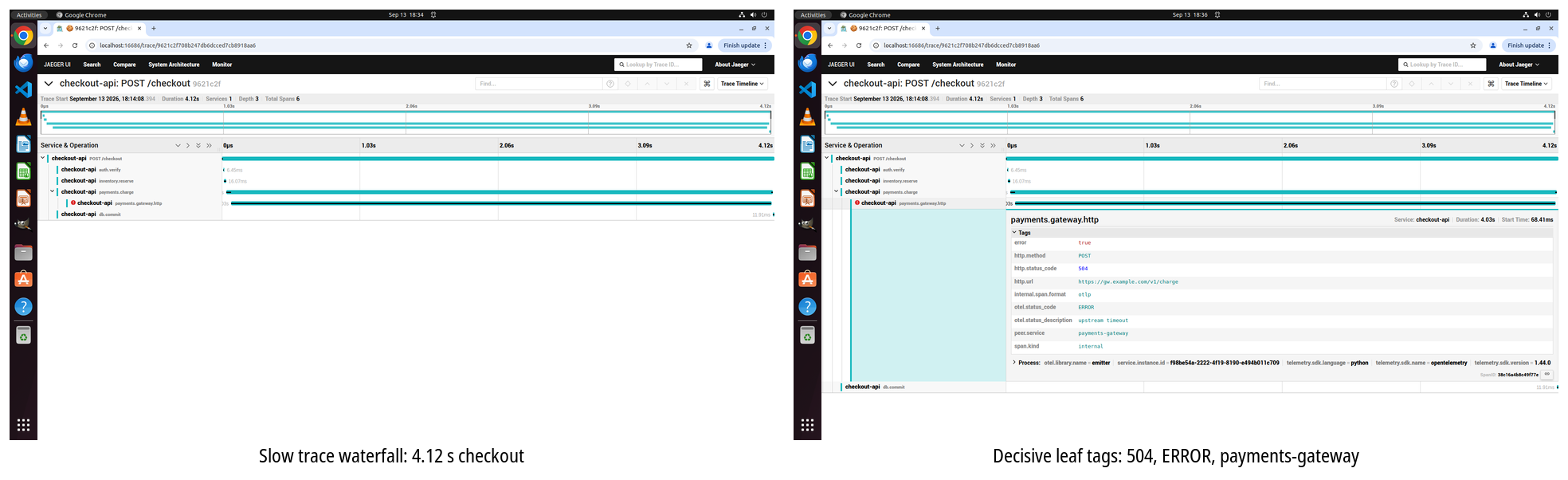}
    \caption{Trace-backed root-cause analysis in WeaveBench. The agent first exports and analyzes the Jaeger traces programmatically, then drives the native UI to verify the causal path visually. The slow-trace waterfall (left) shows \texttt{payments.gateway.http} occupying almost the full 4.12-second checkout request. Expanding that leaf span (right) reads back the decisive evidence: HTTP 504, OpenTelemetry \texttt{ERROR}, an upstream timeout, and \texttt{peer.service=payments-gateway}.}
    \label{fig:weavebench_jaeger_trace}
\end{figure}

\subsubsection{OSWorld-V2: Reconciling a Reimbursement Claim Across Three Web Applications}
\label{sec:appendix:trace:osworld_008}

In task \texttt{008}, the agent had to file a reimbursement claim in
ExpenseFlow for a NeurIPS trip and a Stanford talk. The receipts were in a
MailHub inbox, the charges could be cross-checked in a VaultBank account, and
the claim had to follow a written guideline and the particulars of a previously
submitted report. Instead of opening each e-mail and each bank search in the
GUI, the agent wrote two tools: an 18-line \texttt{mail\_open\_info.py} that
opens an e-mail by its exact subject and prints its body and attachments, and a
12-line \texttt{vault\_search.py} that types a query into the bank's
transaction search and prints the matching rows. It then called them 13 times,
once per supporting e-mail and once per charge, and reconciled every receipt
against a bank posting.





The trajectory used 167 model calls and 163 terminal commands over
approximately 117 minutes and received a final OSWorld-V2 score of
\textbf{0.8816}; the evaluator does not itemize the deduction. As in the
previous traces, the agent verified by readback rather than by trust: it
noticed that the review page showed the expense range as ``02-DEC-2025 --
21-NOV-2025'' because the application sorts date strings lexicographically,
recorded this as a display defect, and left the claim data untouched.

\begin{figure}[ht]
    \centering
    \includegraphics[width=0.95\linewidth]{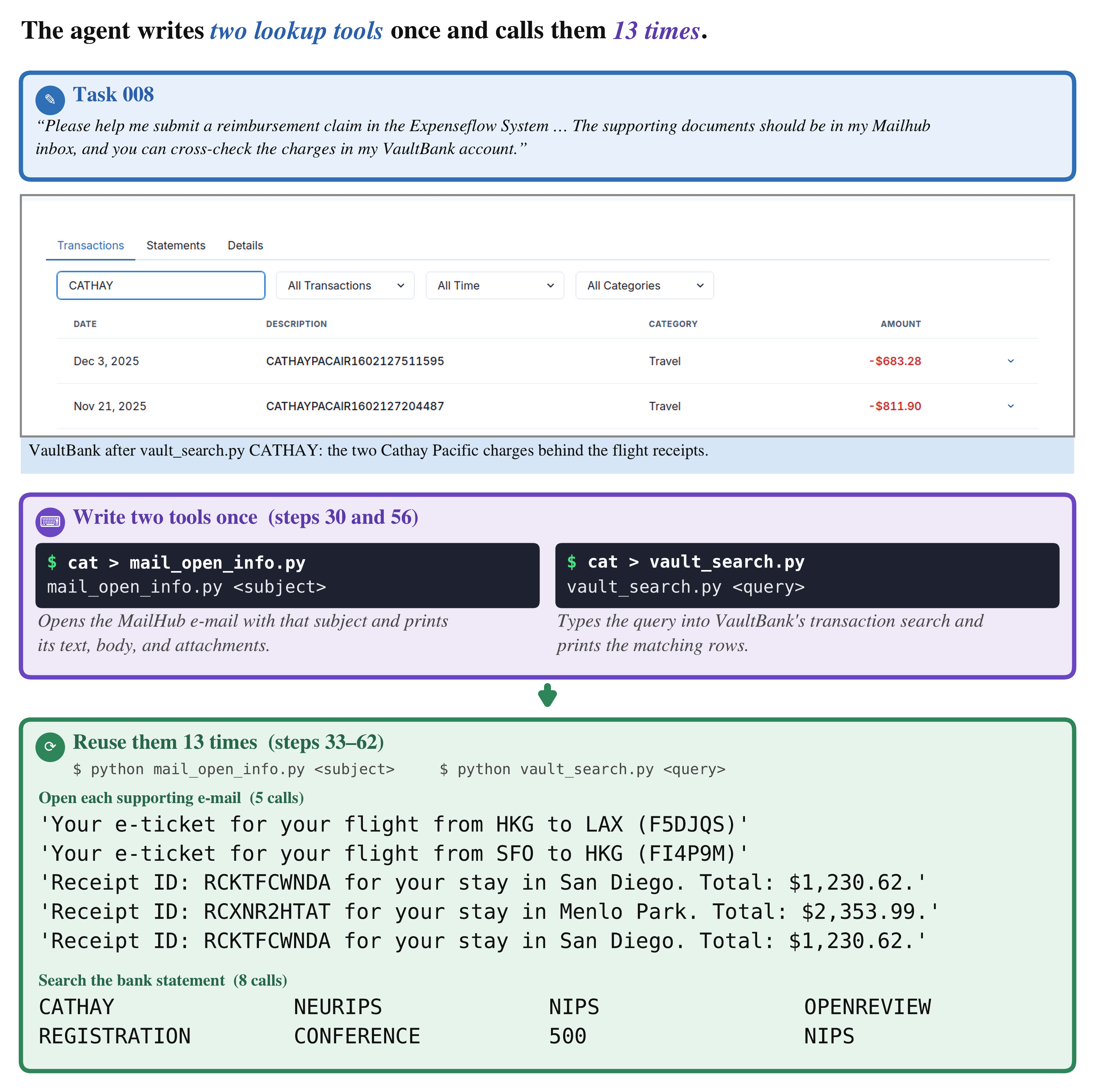}
    \caption{Tool-crafted lookup commands reused across a task in Task008. The task (top) asks for a reimbursement claim whose evidence lives in a mail client and a bank account; the screenshot below it is the bank's transactions tab after the agent's own search tool has typed \texttt{CATHAY}. The agent writes \texttt{mail\_open\_info.py} and \texttt{vault\_search.py} once (middle) and invokes them 13 times with the arguments shown (bottom), once per supporting e-mail and once per charge.}
    \label{fig:tool_reuse_008}
\end{figure}

\subsubsection{OSWorld-V2: Fixing an AWS Billing Anomaly Without a Single Screenshot}
\label{sec:appendix:trace:osworld_070}

In task \texttt{070}, a PI asks in TeamChat for an
abnormal AWS bill to be investigated, fixed in the (mock) console, and
summarized back in the channel. The lab policy allows \$3,000 per month in two
approved regions. The agent switched into the \texttt{lab-admin} role, read the
Billing page (\$11,462 month-to-date against the \$3,000 budget), and found the
cause in \texttt{eu-west-1}: four \texttt{p4d.24xlarge} instances and four
2\,TB \texttt{io2} volumes.

For the remaining console work it wrote a 37-line \texttt{cdp\_click.py} (on
top of a 115-line CDP helper) that resolves an accessibility name, role, or
node id in the live accessibility tree, reads the element's box model, and
clicks its center. It then invoked the script 15 more times, each call a single
line with a new argument, to delete the volumes, deactivate the intern's access
key, and post the incident update.





Every observation in this trajectory was the accessibility tree or page text
printed to stdout. It used 69 model calls and 68 terminal commands over
approximately 20 minutes and received a final OSWorld-V2 score of
\textbf{1.0}.

\begin{figure}[ht]
    \centering
    \includegraphics[width=0.95\linewidth]{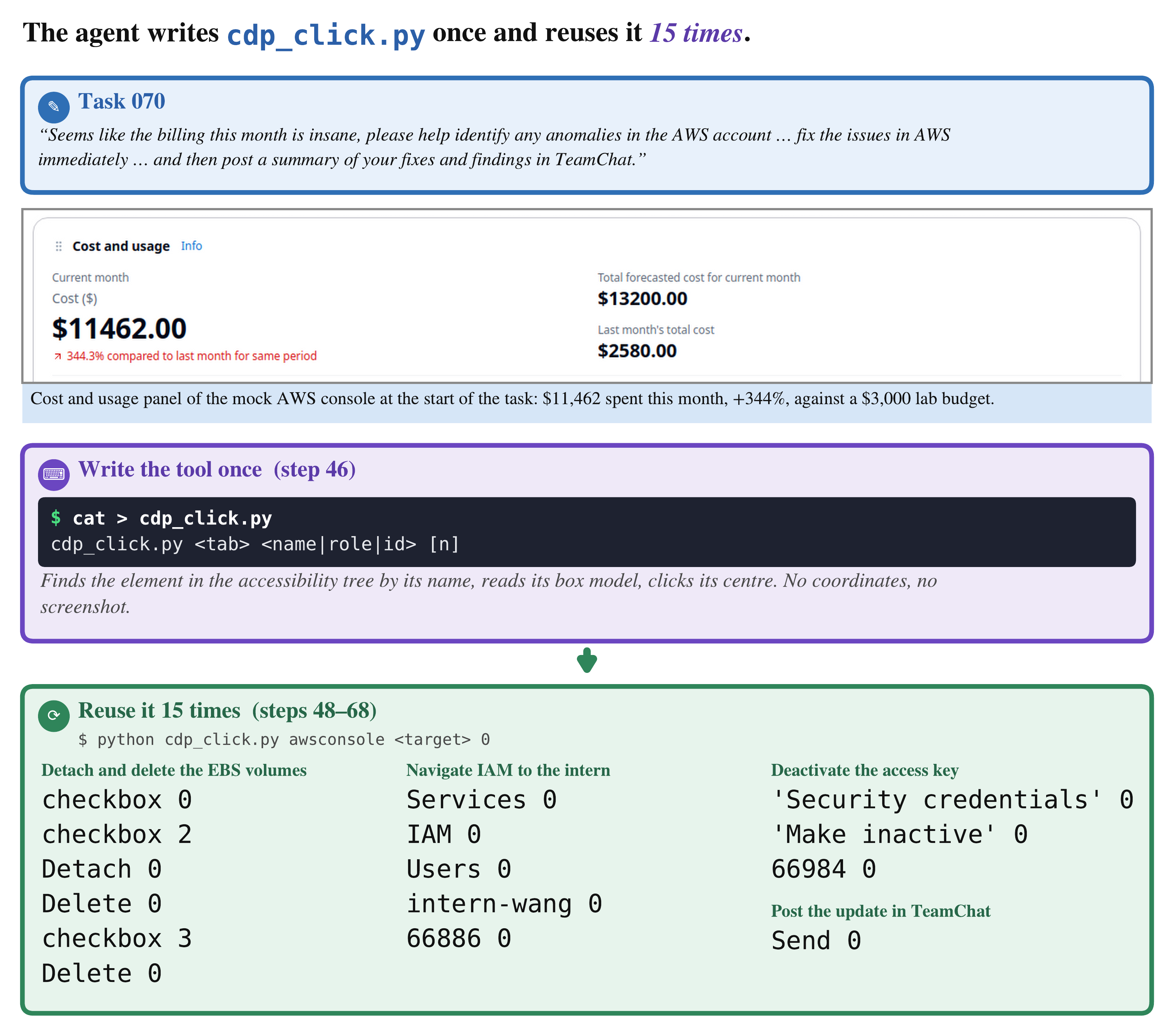}
    \caption{Tool-crafted click primitive reused across a task in Task070. The task (top) asks for an AWS billing anomaly to be found, fixed, and reported in TeamChat; the console's cost panel below it already shows the spike. The agent writes \texttt{cdp\_click.py} once (middle) and invokes it 15 more times with the arguments shown (bottom). The trajectory attaches no screenshot; the console view is taken from the GPT-5.6 run of the same task.}
    \label{fig:tool_reuse_070}
\end{figure}

\subsection{BenchCAD: Reconstructing a Heat Sink}
\label{sec:appendix:trace:benchcad}

We illustrate how CUAWright performs on CAD tasks using the BenchCAD
Vision2Code record \texttt{heat\_sink\_003731\_s20260505}, officially classified
as \texttt{hard} on a scale of easy to hard. The agent received four views of the heat-sink part and had
to write CadQuery code that reproduced its 3D geometry. The submitted STEP file
is scored by voxel intersection-over-union (IoU), from 0 to 1.

CUAWright follows an iterative workflow: write CadQuery code, render the
result, evaluate it against the reference views, and revise the code. This is
analogous to the tool-enabled BenchCAD settings reported by OpenAI and
Anthropic because the model can use execution feedback to improve its answer.
It differs from canonical single-shot BenchCAD, where the model receives one
turn and must return its proposed code without an iterative render-and-repair
loop.

The first model it produced captured the base, seven fins, and four mounting
holes, but gave the fins rectangular tops. After rendering the result, the
agent noticed that each reference fin showed three top edges while its model
showed only two.

The agent tested a triangular roof, a one-sided chamfer, and symmetric chamfers.
Only the symmetric chamfers reproduced the missing edge pattern. It then varied
the fin height, thickness, length, spacing, chamfer size, base thickness, and
hole radius, rendering and inspecting each candidate before choosing the next
change.

The run used 249 model turns and 67 image inspections. The agent executed the
final program, rendered the resulting STEP file, inspected it, and submitted
it. The official BenchCAD scorer returned voxel IoU \textbf{1.0000}. The first
guess was incomplete; execution and visual feedback exposed the missing
feature and enabled the correction.

\begin{figure}[ht]
    \centering
    \includegraphics[width=0.95\linewidth]{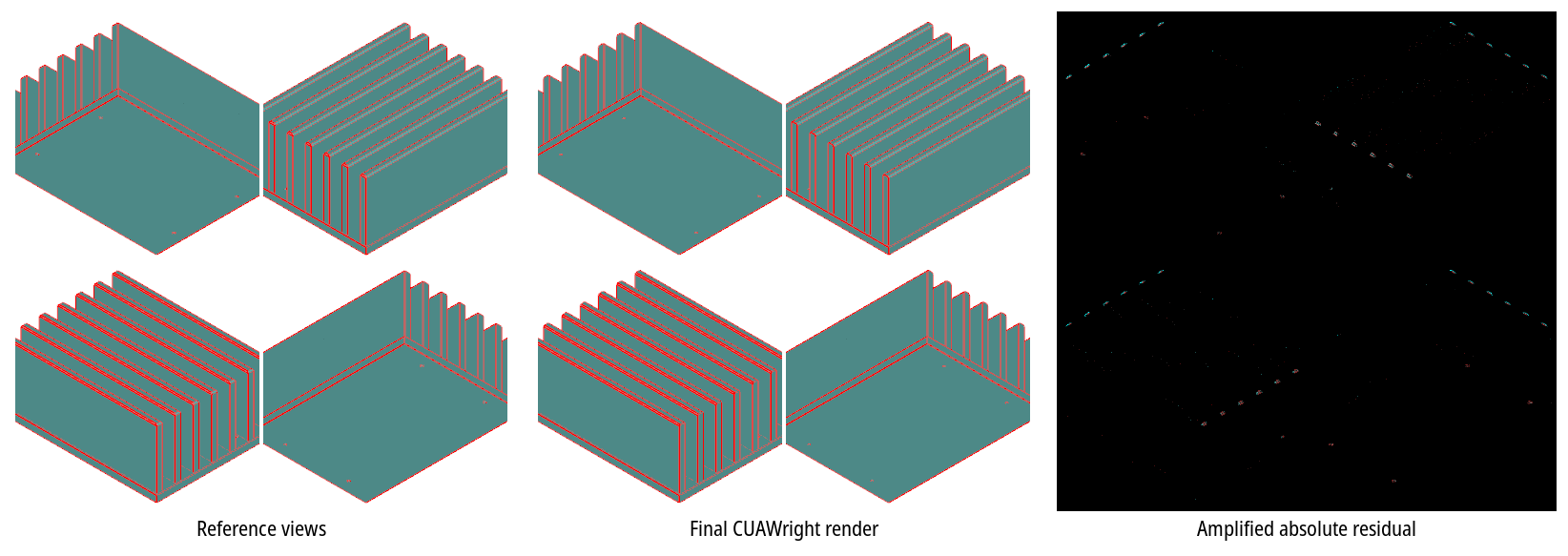}
    \caption{BenchCAD heat-sink reconstruction. Reference views appear on the
    left, the final render in the center, and an amplified pixel difference on
    the right. The submitted STEP file received voxel IoU 1.0000.}
    \label{fig:benchcad_heat_sink_trace}
\end{figure}

\subsection{Direct and Multi-Agent Execution on OSWorld}
\label{sec:appendix:trace:orchestration}

We also examine how CUAWright fares under direct and multi-agent execution on
OSWorld 2.0. Both
conditions used GPT-5.6 Sol at maximum reasoning effort, the same benchmark
release and task evaluator, and a 300-step ceiling.

In direct execution, one Actor owns the complete loop of inspection, editing,
verification, and submission. In multi-agent execution, the Actor shares the
step budget with a Binder that extracts requirements, a Critic that identifies
risks or defects, and a Verifier that decides whether the result is ready to
submit; submission requires its final \texttt{ready} decision. We present two
paired tasks with opposite outcomes: one in which multi-agent review changed
and improved the artifact, and one in which direct execution was more
effective. The examples illustrate how the two modes differ in practice rather
than estimate their average performance.

\begin{table}[ht]
\centering
\caption{\textbf{Paired direct and multi-agent traces.}}
\label{tab:paired_orchestration_traces}
\small
\begin{tabular}{@{}p{0.20\linewidth}p{0.36\linewidth}p{0.36\linewidth}@{}}
\toprule
& \textbf{Task 045: multi-agent advantage}
& \textbf{Task 076: direct advantage} \\
\midrule
Task
& Record all seven dashboard pages in a $1920\times1080$ GIF lasting at least
five seconds and smaller than one MiB
& Apply the speaker-note instructions across six slides, including layout,
sensor, chart, workbook, and table edits \\
Direct
& 0.95; 32 Actor steps; submitted
& 1.00; 42 Actor steps; submitted \\
Multi-agent
& 1.00; 129 total steps; submitted
& 0.95; 300 total steps; not submitted \\
Observed mechanism
& Critic changed the recording method
& Repeated inconclusive verification blocked submission \\
\bottomrule
\end{tabular}
\end{table}

\paragraph{Task 045: multi-agent review improved the artifact.}
The task required a promotional GIF that visited seven dashboard pages at
$1920\times1080$, lasted at least five seconds, showed no obstructing desktop
elements, and remained below one MiB. The direct Actor finished in 32 steps. It
submitted a seven-frame, 8.75-second GIF assembled from selected screenshots.
The artifact covered every page and met the size and resolution requirements,
but received half credit for frame count, for a final score of \textbf{0.95}.

The multi-agent run used 35 Actor, 18 Binder, 12 Critic, and 64 Verifier steps.
The Critic identified that a small set of selected screenshots did not preserve
the actual walkthrough and recommended a genuine screen recording. The Actor
then used \texttt{byzanz-record} while navigating the dashboard. The submitted
GIF contained 156 frames, lasted 10.37 seconds, remained 5,168 bytes below one
MiB, and scored \textbf{1.00}. The report-to-action chain is visible, but the
gain required 97 additional steps.

\paragraph{Task 076: direct execution was more effective.}
The task required six presentation slides to be edited according to their
speaker notes. The requested changes included title formatting, text-box
spacing, movement of four sensor markers, an April chart and embedded-workbook
update, deletion of a temporary table, and alignment of every cell in a budget
table. The direct Actor completed the edits, reopened the file in WPS, checked
the persistent package, submitted in 42 steps, and passed every criterion for
a score of \textbf{1.00}.

The multi-agent run used 198 Actor, 20 Binder, 30 Critic, and 52 Verifier
steps. The Actor completed nearly all requested edits and a final Critic found
no concrete user-facing defect. Three Verifier calls nevertheless returned
\texttt{unknown}. Because submission required a final \texttt{ready} report,
the run exhausted all 300 steps without submitting. Its persistent artifact
scored \textbf{0.95}; the only failed criterion was the Slide 2 body-box gap,
which measured 228,600 EMU against a 990,000-EMU threshold. The direct artifact
measured 1,571,625 EMU.

Task 045 shows where multi-agent review can justify its cost: the Critic
identified a specific artifact risk, the Actor changed its implementation, and
the persistent result improved. Task 076 shows the opposite failure mode:
review repeated without producing a repair and consumed the budget needed for
submission. The traces support Actor-first execution with specialist calls
reserved for named uncertainties that imply a concrete action.

\begin{figure}[ht]
    \centering
    \includegraphics[width=0.98\linewidth]{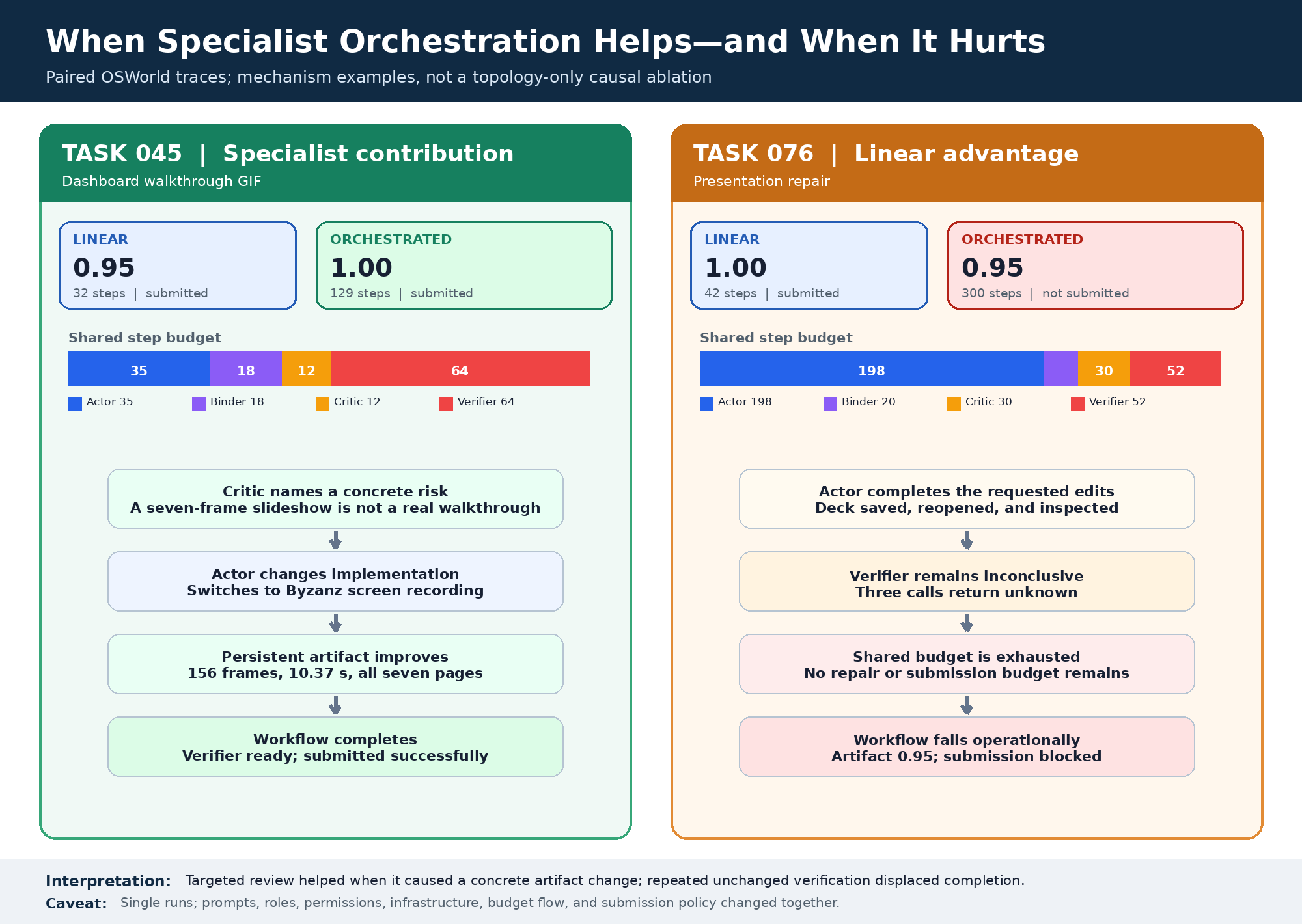}
    \caption{Paired traces for tasks 045 and 076. In task 045, Critic feedback
    changed the artifact from seven selected frames to a 156-frame recording.
    In task 076, three inconclusive Verifier calls exhausted the shared budget
    and blocked submission.}
    \label{fig:osworld_orchestration_paired_traces}
\end{figure}

\end{document}